\pdfoutput=1

\documentclass[11pt]{article}

\usepackage[preprint]{acl}
\usepackage{enumitem}
\usepackage{times}
\usepackage{latexsym}
\usepackage{authblk}
\usepackage[T1]{fontenc}

\usepackage[utf8]{inputenc}

\usepackage{microtype}
\usepackage{algorithm}
\usepackage{algorithmic}
\usepackage{newfloat}
\usepackage{listings}
\usepackage{booktabs} 
\usepackage{multirow}
\usepackage{amsmath}
\usepackage{amssymb}

\usepackage{inconsolata}

\usepackage{graphicx}

\begin{document}

\title{	
From Redundancy to Relevance: Information Flow in LVLMs Across Reasoning Tasks}
\author[1]{\textbf{Xiaofeng Zhang$\dagger$}}
\author[3]{\textbf{Yihao Quan$\dagger$}}
\author[2]{\textbf{Chen Shen}}
\author[2]{\textbf{Xiaosong Yuan}}
\author[2]{\textbf{Shaotian Yan}}
\author[2]{\\\textbf{Liang Xie}}
\author[2]{\textbf{Wenxiao Wang}}
\author[1]{\textbf{Chaochen Gu*}}
\author[4]{\textbf{Hao Tang}}
\author[2]{\textbf{Jieping Ye}}

\affil[1]{Shanghai Jiao Tong University}
\affil[2]{Alibaba Group}
\affil[3]{Beijing Jiaotong University}
\affil[4]{Carnegie Mellon University}
\affil[ ]{\{framebreak@sjtu.edu.cn\}}
\maketitle


\begin{abstract}
Large Vision Language Models (LVLMs) achieve great performance on visual-language reasoning tasks, however, the black-box nature of LVLMs hinders in-depth research on the reasoning mechanism. As all images need to be converted into image tokens to fit the input format of large language models (LLMs) along with natural language prompts, sequential visual representation is essential to the performance of LVLMs, and the information flow analysis approach can be an effective tool for determining interactions between these representations. In this paper, we propose integrating attention analysis with LLaVA-CAM, concretely, attention scores highlight relevant regions during forward propagation, while LLaVA-CAM captures gradient changes through backward propagation, revealing key image features. By exploring the information flow from the perspective of visual representation contribution, we observe that it tends to converge in shallow layers but diversify in deeper layers. To validate our analysis, we conduct comprehensive experiments with truncation strategies across various LVLMs for visual question answering and image captioning tasks, and experimental results not only verify our hypothesis but also reveal a consistent pattern of information flow convergence in the corresponding layers, and the information flow cliff layer will be different due to different contexts. The paper's source code can be accessed from \url{https://github.com/zhangbaijin/From-Redundancy-to-Relevance}

\end{abstract}


\section{Introduction}

Multimodal models are more prevalent due to the capability of understanding vision-language rather than merely textual information. Most large vision-language models (LVLMs) consist of various visual encoders and large language models (LLMs). When images are fed to LLMs together with language prompts, they are transformed into hundreds or thousands of tokens \cite{LLaVA,bai2023qwen,shikra,zhu2024llava,minigemini,instructblip,minigpt4,internvl,mplug,mobilevlm,video-llama,imagebind,next-gpt,cogvlm,minigpt2}. Such a transformation leads LVLMs to rely on sequential visual representation. Albeit popular LVLMs exhibit impressive generation capabilities, the black-box design of Transformers decoder-stacked LLMs hinders the interpretability of visual-language models. In this paper, we intend to explore the inner mechanisms during LVLMs reasoning.

\begin{figure*}[h]
\centerline{\includegraphics[scale=0.22]{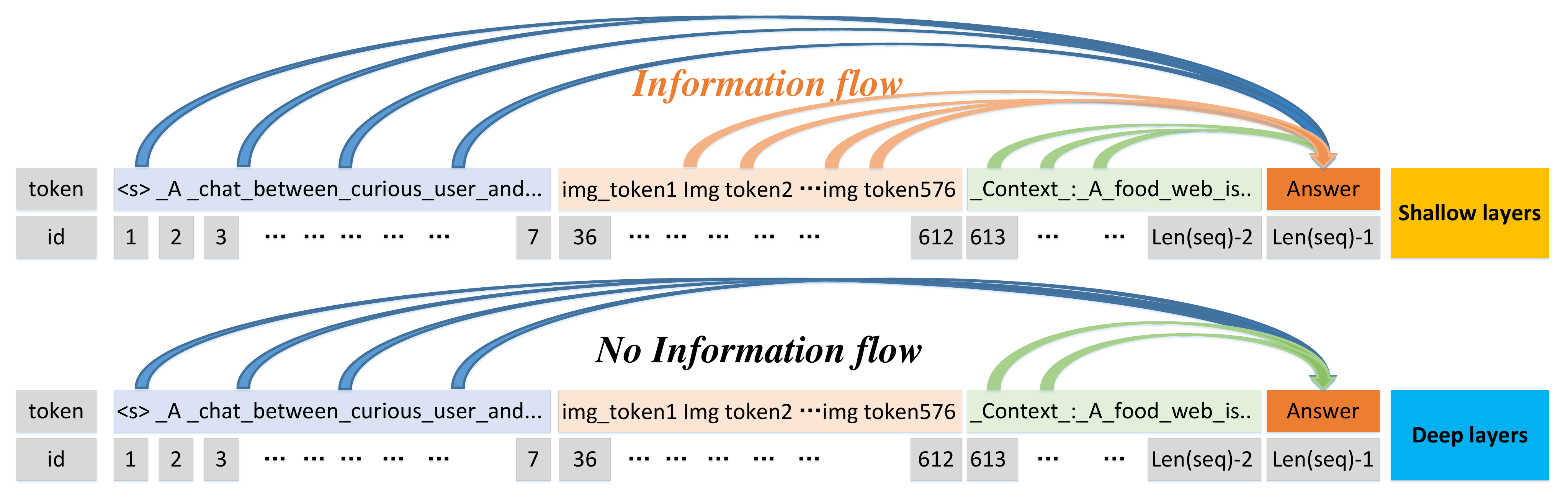}}
\caption{It shows the information flow of tokens, from left to right are system tokens, image tokens, user tokens, and output tokens. There is a convergence of the information flow of the system token, image token, and user token towards the output token at the shallow layers. The convergence of the information flow of the system token and user token is much more obvious than the image token at the deep layers, which we can call the deep layers as information flow cliff layers.}
\label{information-flow}
\vspace{-0.4cm}
\end{figure*}

Prior works have commenced preliminary exploration into the caption mechanisms of LLMs~\cite{label-words,highlighter,function,knowledge}. OPERA~\cite{opera} traces the potential causes of hallucinations in LVLMs via attention maps, suggesting that during the inference phase, the model may produce hallucinations by sequentially summarizing previous tokens when processing special tokens such as `-', `?', and other special symbols. 
OPERA mitigates hallucinations by imposing penalty constraints on attention scores, marking the first work to visualize multimodal hallucinations.
In addition, FastV~\cite{fastv} identifies that the computation of attention for visual tokens in the deep layers (near the output) of the LVLMs is extremely inefficient. Although previous studies have made significant strides in enhancing LVLMs perception by leveraging attention-based approaches, they focus less on the dynamic interactions between images and texts. To this end, we aim to deepen the understanding of how images and texts influence each other within reasoning tasks. We define \textbf{`information flow'} as the degree of influence of image, user, and system tokens on answer tokens. Understanding the information flow between image and text tokens is crucial for dissecting multimodal reasoning.

\begin{figure}[h]
\centerline{\includegraphics[scale=0.15]{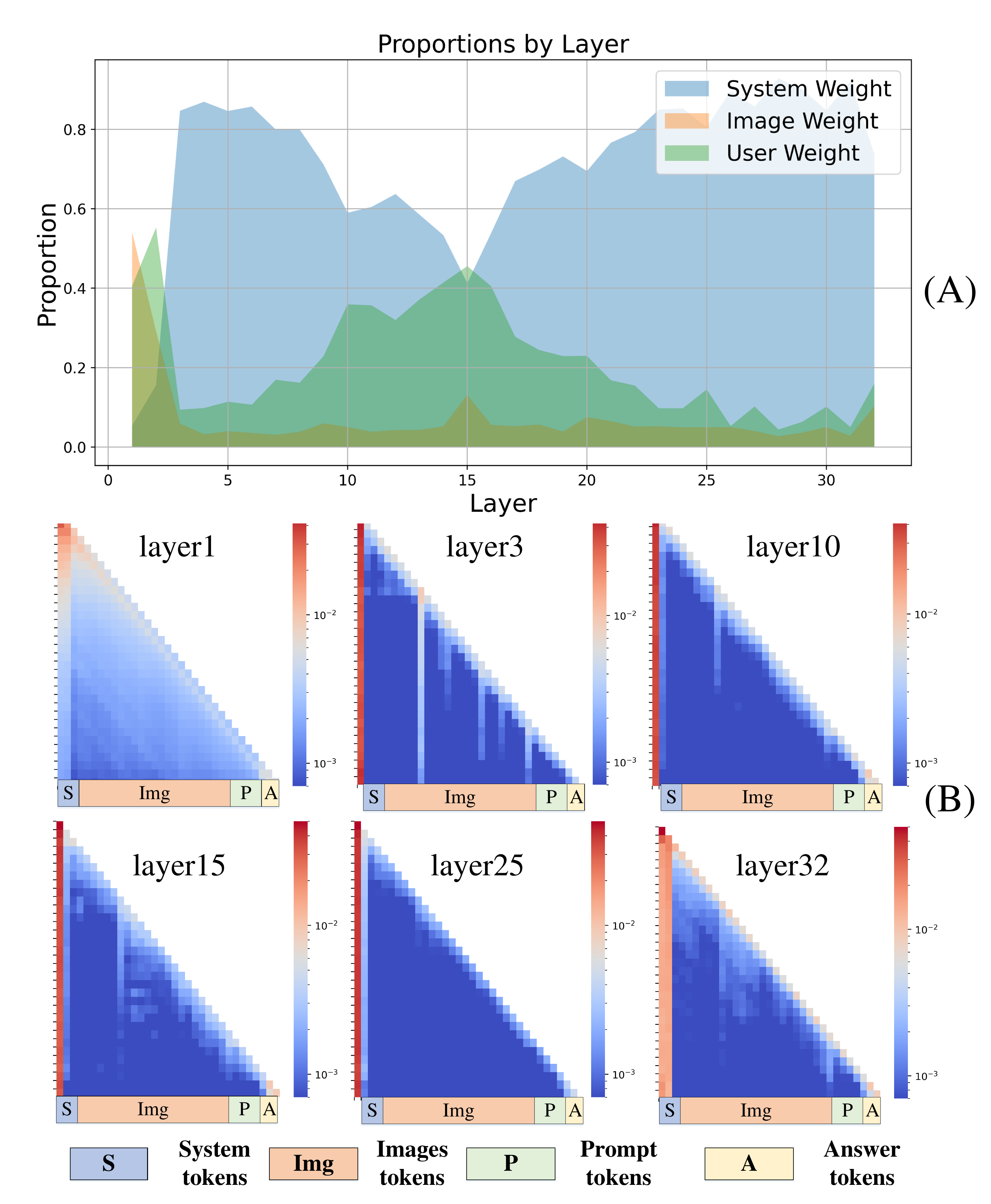}}
\caption{(A) is the percentage of system tokens, image tokens, and prompt tokens on the attention weight of the answer. (B) is the attention map of system tokens, image tokens, prompt tokens, and answer tokens. It can be observed that the attention scores for image tokens decrease rapidly in layers 1-5 and stabilize in layers 6-31. The attention allocated to image tokens is significantly lower throughout these layers than system and user tokens. However, image and user tokens' attention scores increase rapidly at the 32nd layer.}
\label{weight-results}
\end{figure}

In this paper, as shown in Figure~\ref{information-flow}, we discover an important phenomenon in a view of information flow: LVLMs tend to depend on the prompt, image tokens do not impact the answers in the middle or deep layers. We refer to this first sparse layer as the information flow cliff layer. As shown in Figure~\ref{weight-results}, we first visualize the attention scores and maps for system, image, and prompt tokens in relation to the answer. We observe that after the third layer, image tokens account for less than 2$\%$ of the total compared to system and prompt tokens, with their weights ranging from 0 to 0.2. This raises two questions: 

\textbf{Q1:} \textit{Is there any information flow from the third layer to the deeper layers?}

\textbf{Q2:} \textit{Does the attention score give a complete reflection of the information flow of the image token?}

Though attention scores highlight relevant regions during forward propagation, they don't fully explain the role of image tokens as they lack gradients and can't reveal the model's decision-making process. To address this, we introduce LLaVA-CAM, which captures gradient changes through backward propagation to show how image features contribute to the answer. With LLaVA-CAM, we consistently observe that image token information flow is concentrated in the shallow layers due to differences in the prompt, while deeper image tokens contribute little to the answer. We refer to the deeper layer where the image ceases to contribute to the answer as the \textbf{`image information flow (i-f) cliff layer'}.



Given the information flow convergence in shallow layers, we find that the LLaVA-CAM results vary in distinct tasks. For example, in complex reasoning datasets like ScienceQA~\cite{sqa}, the model focuses on relevant regions in shallow rather than deeper layers. In contrast, the model emphasizes relevant regions from shallow to deep layers for general reasoning tasks such as TextVQA~\cite{textvqa} and POPE~\cite{pope}. To validate this hypothesis, we conduct multiple truncation experiments. Empirical results show that when applying fully truncated image tokens based on attention scores, certain layers can maintain the model's inference accuracy and even enhance performance in some cases. Our multi-perspective analysis elucidates the model's inner mechanism and reveals the complex interactions during inference. Notably, we verify in the models LLaVA~\cite{LLaVA}, Intern-VL~\cite{internvl}, Qwen-VL~\cite{bai2023qwen}, Shirak \cite{shikra}, InstructbLIP2 \cite{instructblip} and LLaVA1.6 \cite{llava1.6} that the information flow converges in the shallow or middle layers, and becomes sparse in the deeper layers, and that there are information-flow cliff layers in each of the models, this means that image tokens are highly redundant after the cliff layer. Our methods offer a new perspective for interpretability in LVLMs. In summary, our main contributions are as follows:



\setlist[itemize]{left=0em}
\begin{itemize}
\item We propose a novel information flow analysis method, which combines LLaVA-CAM and attention score to explore the interaction mechanisms in LVLMs reasoning tasks. 
\item We find a prevailing pattern in multiple LVLMs: the information flow converges in shallow layers while diverging in deeper layers. This means that image tokens are highly redundant after the cliff layer.
\item Results of truncation experiments validate our observation of information convergence in the corresponding layers across various models.
\end{itemize}

\begin{figure*}[t]
\centerline{\includegraphics[scale=0.28]{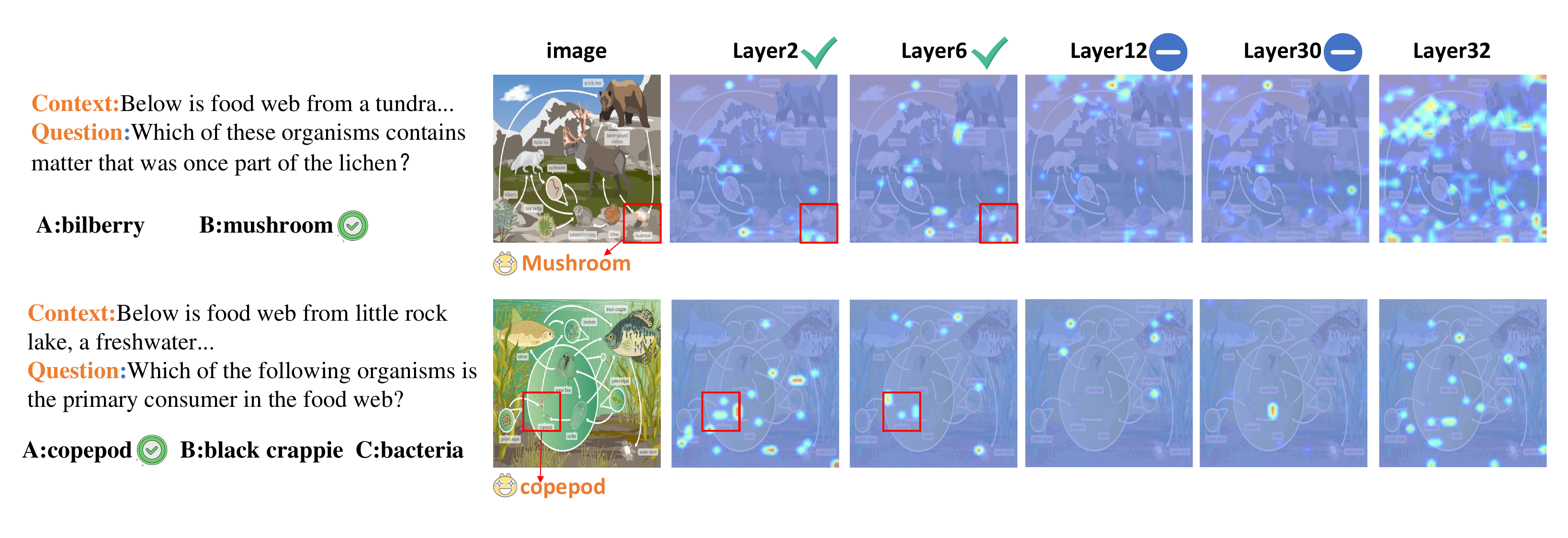}}
\caption{The LLaVA-CAM results of LLM on ScienceQA dataset(Complex reasoning). The information flow of the image converges to the correct region in the early layers and diverges in the deeper layers, and then the information flow cliff layer begins to appear.}
\label{grad-llm}
\vspace{-0.2cm}
\end{figure*}

\section{Related work}

\subsection{Information Flow and Interpretability in LVLMs}


Recent work by Wang et al. \cite{label-words} has made significant strides in understanding information flow through the concept of label words as anchors. This approach focuses on how information aggregates, using a combination of attention scores and gradients to measure the saliency of specific tokens. In their attention-score matrix, the convergence of key tokens is evident along the ordinate axis. Label words will converge on the user prompt first in the Zero-shot ICL/CoT case. In their findings, Wang et al. observed that each shot will converge on the last token in the few-shot ICL/CoT case. This insight is crucial for analyzing truncation strategies in models for Visual Question Answering (VQA) and image captioning tasks. OPERA \cite{opera} introduces a novel method for identifying hallucinations in LVLMs by examining attention maps. It detects hallucinations during the sequential summarization of tokens following key symbols such as ‘-’ and ‘?’. By imposing penalty constraints on attention scores, OPERA pioneers the visualization of multimodal hallucinations. Complementing this, DOPRA \cite{DOPRA} investigates the information flow of the output of each layer of the transformer.

Expanding the scope further, ACT \cite{attention-sink} investigates attention sinks within Large Language Models (LLMs) and seeks to improve model accuracy by optimizing attention distributions directly, without requiring fine-tuning. This work begins with detailed visualizations of attention distributions across various inputs and tasks during inference, intending to enhance our understanding of attention mechanisms in LLMs.

Similarly, FastV \cite{fastv} identifies inefficiencies in attention mechanisms within LVLMs, particularly in models such as LLaVA-1.5 \cite{LLaVA}, QwenVL \cite{bai2023qwen}. The authors observed that the computation of attention for visual tokens in the deep layers of these models is extremely inefficient. To address this, they proposed an image-token pruning strategy at specific layers, aiming for a sparser approach than textual data processing.

Although previous studies have made significant strides in enhancing LVLMs' perception and inference speed through leveraging attention-based mechanisms, they tend to focus less on the dynamic interplay of image-text interactions. Recognizing this gap, our research aims to deepen the understanding of how images and texts influence each other within complex reasoning. 

\section{The Proposed Method}
To visualize the information flow, LLaVA-CAM, and attention score are used to provide a comprehensive view of the information flow. Attention scores illuminate the model's forward propagation, while LLaVA-CAM delves into the backward propagation, reflecting how various input elements contribute to the final prediction. This dual-method approach not only quantifies the significance of each input element but also provides a visual representation of their influence on the model's predictive decisions.

\subsection{ LLaVA-CAM for Large Vision Language Models}

Our method is inspired by Smooth-CAM \cite{smooth}. Complex reasoning is still essentially a text generation task, and the answer generation is a sentence made up of the classification results for each word. The network outputs probabilities for the $n$ tokens denoted as $z = [z_i, .. z_n]$,
to visualize the model's output answer using LLaVA-CAM:
\begin{equation}
z_{\mathrm{answer}} = \sum_i^n z_i,
\end{equation}
to obtain the logits representing the overall output of the model. The bias $G_k$ is applied to all the feature maps $A_{k}$ of the layer output by $z$:


\begin{equation} 
G_k=\frac{\partial z_c}{\partial A_k} ,
\end{equation} 
all attention mappings $A_k$ in the last layer of the image encoder/LLM decoder, solve for $z_{\mathrm{answer}}$, where $A_k$ represents the feature map at the coordinate point of the $n^{\mathrm{th}}$ channel, $\alpha$ is the weight vector, and the resultant derivative feature map $G$, $c=1,2,3..n$. Then transform the sequence inputs to the two-dimensional shape of the H$\times$W. For sequence inputs, identify the sequence ID associated with image tokens. For instance, in LLaVA1.5, id (35-611) corresponds to image tokens. Then, multiply the $\alpha$ weight vector by the relevant channels of the feature map $A$ to generate a two-dimensional activation map:
\begin{equation}
Heat_{\mathrm{cam}} = \operatorname{ReLU}(\sum_k \alpha_k A_k),
\end{equation}
where the ReLU function creates a heat map highlighting the positive correlation between activation values and the correct class.
Next, Gaussian noise is added to the image to generate multiple perturbation samples:
\begin{equation}
\mathbf{x}_{\text {noisy }}^{(i)}=\mathbf{x}+\mathbf{N}\left(\mu=0, \sigma^2=noise_{s}\right),
\end{equation}
where $noise_{s}$ represents the standard deviation of the Gaussian noise added to the input image and $i$ represents the first $i$ generation of the image with noise. The CAM plots are averaged by generating the perturbation samples multiple times:
\begin{equation}
 \text {LLaVA-CAM}=\frac{1}{N} \sum_{i=1}^N Heat_{\mathrm{cam}}\left(\mathbf{x}_{\text {noisy }}^{(i)}\right).
\end{equation}

Finally, we apply maximum normalization to the cam map. The heat map is then color-mapped and overlaid on the original image, providing a clear visual representation of the model’s attention distribution.




\begin{figure*}[h]
\centerline{\includegraphics[scale=0.32]{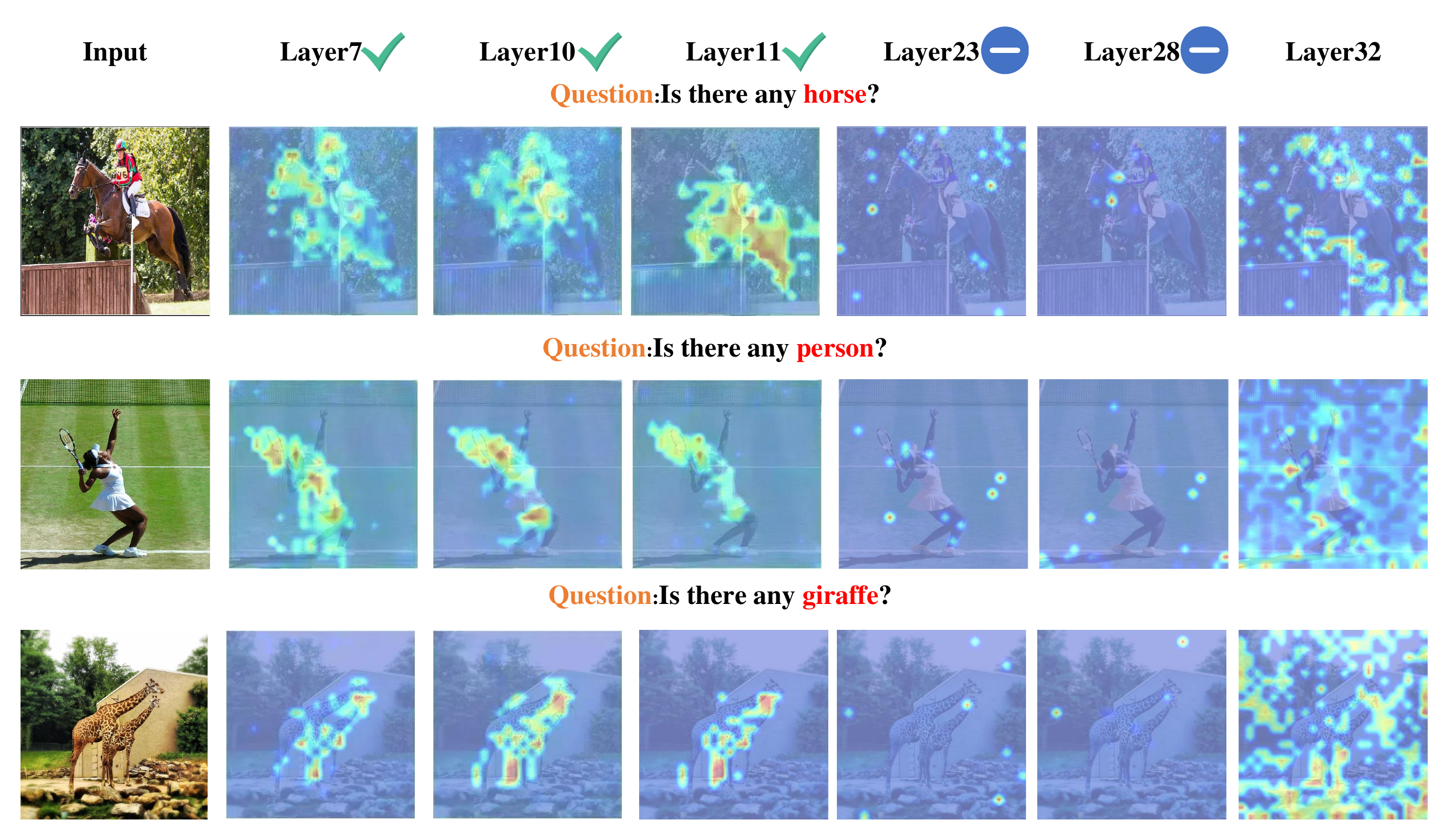}}
\caption{The LLaVA-CAM results of POPE \cite{pope} and TextVQA \cite{textvqa} (Common reasoning). It can be analyzed from the LLaVA-CAM diagram of VQA that when the model recognizes the confirmed recognition objects such as "horse, people", etc., It will focus on the corresponding areas from the first layer to the deeper layers until a cliff layer occurs and causes information flow to be sparse.}
\label{Smooth-CAM3}
\end{figure*}

\subsection{LLaVA-CAM for Exploring Information Flow in LLM}

The abundant features from the encoder are passed into the LLM decoder which interacts with the text. We use LLaVA-CAM to visualize the information flow of image tokens and their impact on answer tokens in the LLM decoder. As shown in Figure~\ref{grad-llm} and Figure~\ref{Smooth-CAM3}.

\textbf{Pattern:} In tasks involving complex reasoning, common reasoning, or image captioning, information flow converges in the early or middle layers and starts to disperse in the deeper layers. It is plausible that the 32nd layer functions as the retrospective layer for deep reasoning, potentially integrating both the image and user prompt. 

\begin{figure*}[!t]
\centerline{\includegraphics[scale=0.25]{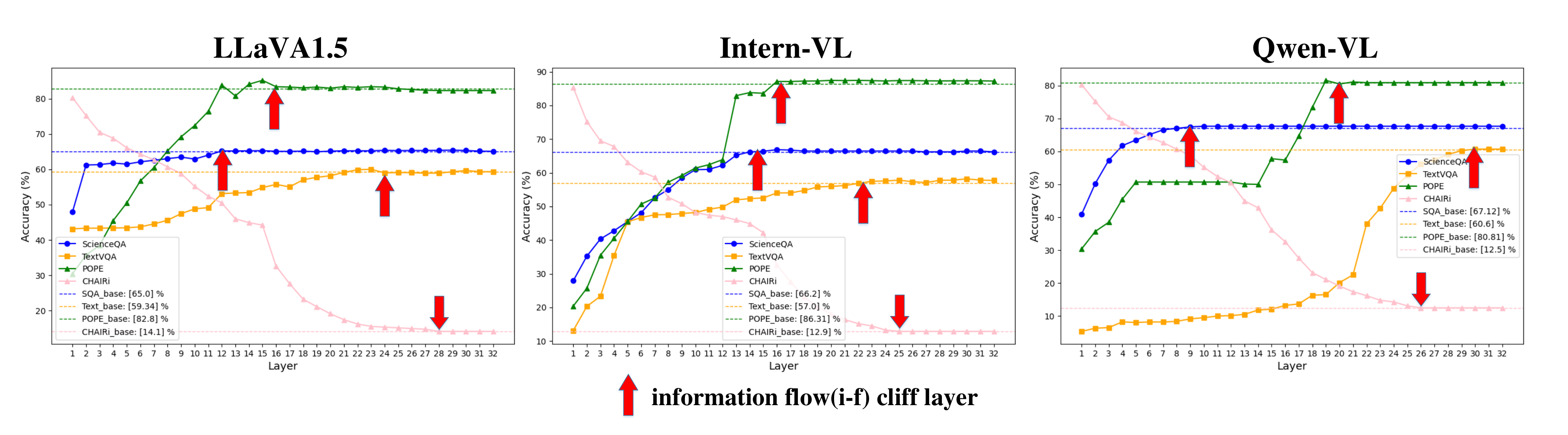}}
\caption{The truncating 576 image tokens experiments on three VQA datasets include POPE/TextVQA/ScienceQA/ and a caption dataset CHAIR dataset, where the red arrow represents the information flow cliff layer. LLaVA1.5-7B \cite{LLaVA}, Intern-VL 7B \cite{internvl}, and Qwen-VL 7B \cite{bai2023qwen} all conform to the pattern of information flow convergence in the early layer and dispersion in the deep layer. Deeper layers can exhibit cliff layers, where truncating image tokens no longer affects the model's accuracy. }
\label{token0}
\end{figure*}

(1) \textbf{Complex reasoning:} In the context of complex reasoning tasks like ScienceQA \cite{sqa}, when the prompt explicitly references specific regions within an image, such as those depicted in the first two lines of Figure~\ref{grad-llm} (\textit{mushroom, copepod}), the model swiftly converges on the relevant areas in the early layers. The deeper layers appear to serve as the foundation for deep reasoning, with the complexity of the prompt dictating the layer at which information dispersal initiates, as illustrated in Figure~\ref{token0}, the information flow cliff layer appears at layer 12 (LLaVA1.5).

(2) \textbf{Common reasoning:} For common reasoning tasks such as TextVQA \cite{textvqa} and POPE \cite{pope}, the model consistently allocates attention to the key regions from the first layer to the deep layers. This consistency may suggest that fewer layers are necessary for deep reasoning, leading to the information flow diverging at deeper layers, specifically at the 18th and 24th layers for TextVQA and POPE, respectively, as indicated (A) in Figure~ \ref{token0}. Furthermore, simpler tasks like image captioning, which require sustained attention to global image features, lead to a delayed dispersion of information flow and consequently require fewer layers for reasoning.



\subsection{Attention-score for Exploring Information Flow in LLM}
\label{attention-score-llm}

\begin{figure}[!t]
\centerline{\includegraphics[scale=0.38]{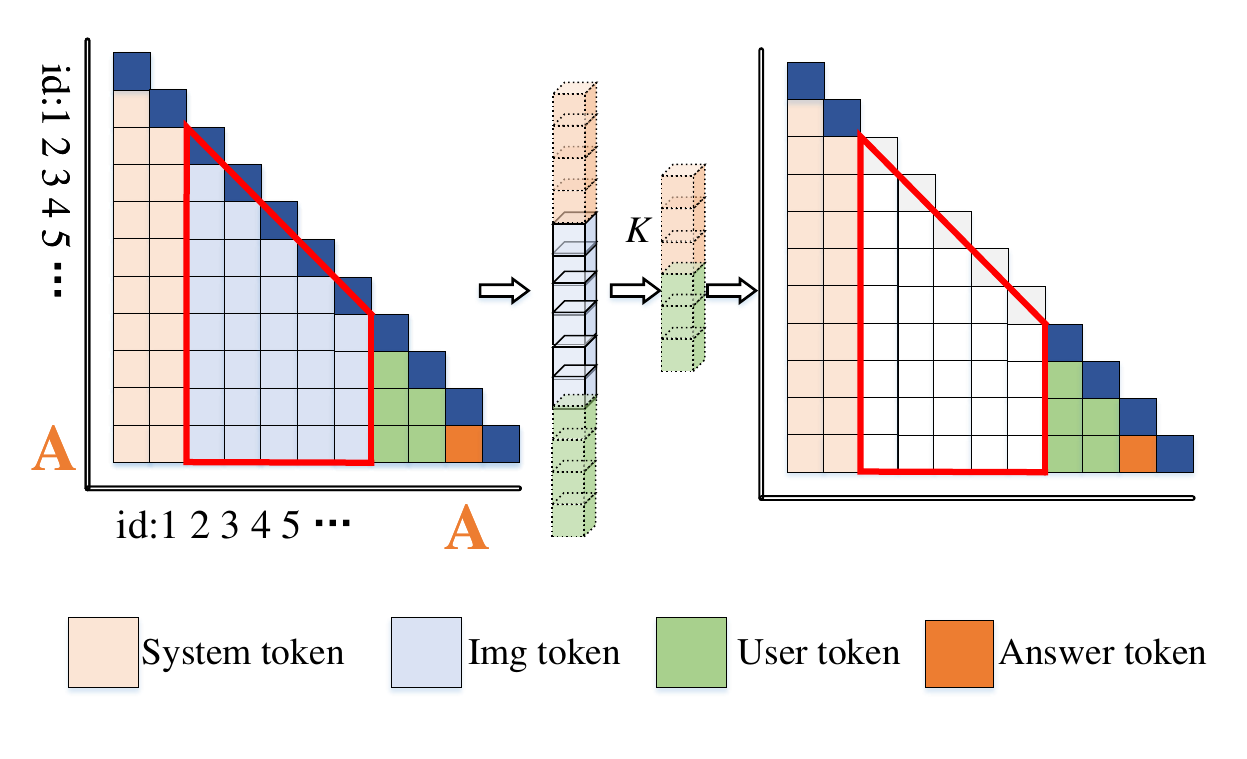}}
\caption{ The diagram of attention-score truncation, $A$ represents the answer token id index, $K$ is set 0.}
\label{structure-turtune}
\vspace{-0.4cm}
\end{figure}

In the above section, we employ LLaVA-CAM to visualize the contribution of the image token to the output token. We found that the models converge in the shallow and middle layers and diverge in the deep layer. In this section, as shown in Figure~\ref{weight-results}, to validate the observed phenomenon of information flow convergence highlighted in LLaVA-CAM, we employ attention-score computation to help us understand the model’s reasoning mechanisms, clarifying their contribution to the final output. We can denote the attention score of all the tokens in the output token as the influence rate, denoted by $w$. This measure can be aggregated for different types of input tokens.
For the output token of the reasoning task, in the $n$-th layer, we define $\mathcal{G}$ as the indices set of all tokens and $\mathcal{G}$ can be divided into three parts that represent the indices set of system, image, and user tokens:
\begin{equation}
\label{jihe}
    \mathcal{G} = \mathcal{S} + \mathcal{I} + \mathcal{U},
\end{equation}
where $\mathcal{S} = \{ 1, \dots, N_{\mathrm{sys}} \}$ represents the index of system token, $N_{\mathrm{sys}}$ represents the length of system token, $\mathcal{I} = \{ N_{\mathrm{sys}} + 1, \dots, N_{\mathrm{sys}} + N_{\mathrm{img}} \}$ represents the index of image token, $N_{\mathrm{img}}$ represents the length of image token, and $\mathcal{U} = \{ N_{\mathrm{sys}} + N_{\mathrm{img}} + 1, \dots, N_{\mathrm{sys}} + N_{\mathrm{img}} + N_{\mathrm{user}} \}$ represents the index of user token, $N_{\mathrm{user}}$ represents the length of user token. $A_{i,j}$ is defined as the total attention score of the output token's attention on different types of tokens. For the $i$-th query token, the attentions from system, image, and user tokens are summed as $1$:
\begin{equation}
 \label{eq1}
\sum_{j \in {\mathcal{S}}}A_{i, j}+\sum_{j \in {\mathcal{I}}}A_{i, j}+\sum_{j \in {\mathcal{U}}}A_{i, j}=1,
\end{equation}
to ensure that the sum of attention scores for each token is 1, it is necessary to normalize the above summation results to calculate the total attention score for the image token:
\begin{equation}
\lambda_{\mathrm{img}}^j=\sum_{j \in {\mathcal{I}}}A_{i, j} .
\end{equation}
There are 576 image tokens in LLaVA1.5 and 256 image tokens in Qwen-VL \cite{bai2023qwen}. From Figure~ \ref{weight-results}, we can see that image tokens account for most of the input tokens, but they receive significantly less attention; instead, the system prompt, which provides the least semantic information, attracts the most attention scores. 

\textbf{Pattern:} Figure~\ref{weight-results} and Figure~\ref{grad-llm} represent the attention score and LLaVA-CAM, respectively, and they show the same pattern, the information flow of image tokens converged in the shallow layers and dispersed in the deep layers. Notably, at layer 32, the attention score of the prompt and image tokens increases, and LLaVA-CAM shows that a large number of features contribute to this layer. This might indicate that layer 32 acts as a retrospective layer and the model potentially refocuses on the image and prompts before making its final decision.


\subsection{Image Token Truncation by Attention-score}
\label{explore-truncation}

As discussed previously, our observations indicate that the information flow of image tokens converges at shallow layers and diverges at deeper layers. To verify whether the information flow diverges at deeper layers, we design a truncation strategy for image tokens as shown in Figure~ \ref{structure-turtune}. In the attention matrix, each row represents the attention score of a query token to all key tokens. 
Let $H$ denote the total number of attention heads, and let \(S\) denote the sequence length. The computation of the average attention matrix \(A\) across the different heads for the attention map \(O\) at the $l_{th}$ layer is formulated as follows:
\begin{equation}
    A = \frac{1}{H} \sum_{h=1}^{H} O_h.
\end{equation}
From this matrix, we identify the indices of tokens with the highest attention scores within specific ranges, defining the corresponding attention matrix segments as:

\begin{equation}
    A_{\mathrm{img}} = A_{N_{\mathrm{sys}} + N_{\mathrm{img}}, \mathcal{I}},
\end{equation}
where $A_{\mathrm{img}}$represents the attention matrix segment for image tokens.
Further, we define:
\begin{equation}
A_{i, \mathcal{K}} = [ A_{i, k_{1}},  A_{i, k_{2}}, \cdots ] \in \mathbb{R}^{\vert \mathcal{K} \vert},
\end{equation}
where$\qquad \mathcal{K} = \{k_{1}, k_{2}, \cdots \} $ and $A_{i, \mathcal{K}} \in \mathbb{R}^{\vert \mathcal{K} \vert}$ represents the vector composed of elements whose row index is $i$ and column indices belonging to the set $\mathcal{K}$. Here, $\vert \mathcal{K} \vert$ denotes the number of elements in the set $\mathcal{K}$.

To pinpoint the top $\mathcal{K}$ contributing elements, we utilize the function $\operatorname{argtop}(\cdot, k)$, which retrieves the indices of the top $\mathcal{K}$ elements. To validate the information-flow cliff layers, here $k$ is set to 0.

\begin{equation}
\mathcal{I}' = N_{\mathrm{sys}} + \operatorname{argtop}(A_{\text{img}}, k).
\end{equation}

Finally, we combine the above formula with Eq. \eqref{jihe} to obtain:
\begin{equation}
    \mathcal{G}' = \mathcal{S} + \mathcal{I}' + \mathcal{U}.
\end{equation}

\section{Experiment}

\subsection{Dataset and Implementation Details}
The ScienceQA \cite{sqa} is the Chain-of-thought dataset for complex reasoning, containing 21,208 multimodal questions (both visual and textual), along with corresponding answers, background knowledge (lectures), and explanations. 
POPE \cite{pope} evaluates hallucination by having the model answer true/false questions about the presence of objects in an image (e.g., `Is there a car in the image?'). TextVQA \cite{textvqa} involves identifying text-related questions, detecting text areas, converting image text to text representations, focusing on relevant text areas, and determining if the final answer requires reprocessing. CHAIR \cite{chair} metric is utilized to assess the phenomenon of object hallucination in the domain of image captioning. It quantifies the ratio of objects referenced in the caption that are not present in the actual image. This metric is available in two forms: CHAIR$_{I}$, which evaluates hallucination at the individual object instance level, and CHAIR$_{S}$, which does so at the sentence level.


Our experiments were conducted on an A100 GPU. It should be noted that we utilize $model$ in the replication of the model, defaulting to greedy search to avoid interference from other parameters.


\subsection{Truncation Experiments for Verifying Information Flow Cliff Layer}

\label{truncation}
As shown in Figure~\ref{token0}, we introduce an early truncation strategy to investigate the information flow of image tokens in shallow layers. This strategy involves truncating all the image tokens to prevent the information flow within LLMs.



When handling complex reasoning tasks such as ScienceQA \cite{sqa}, which present lengthy and complex prompts with multiple choice options, the model's image tokens tend to converge on relevant regions in shallow layers. As shown in Figure~\ref{token0}, starting from layer 12, the information flow begins to diverge. It seems that more layers might be necessary for deep reasoning in intricate tasks. This early divergence could also suggest a possible need for deeper processing in complex reasoning tasks. In contrast, for simpler tasks such as TextVQA and POPE, the model maintains its focus on the correct reasoning regions from the shallow to the deep layers. Divergence in the information flow is observed at layers 18 and 22, respectively, indicating that fewer layers are needed for deep reasoning in these tasks.

Compared to the ScienceQA and TextVQA tasks, image captioning is less complex, requiring the model to focus on the global features of the image. It seems that the requirement for deep reasoning layers is comparatively reduced, with the divergence of information flow occurring at later layers, such as around layer 30 in LLaVA1.5.

Therefore, we can conclude that a consistent pattern emerges in these different types of datasets: the information flow converges at shallow layers and diverges at deeper layers. One possible explanation is that the more complex the reasoning task is, the more layers of reasoning are needed, which means that the layer where the information flow diverges appears more forward. The 32nd layer seems to be like a retrospective layer; after multiple layers of thought and reasoning, the retrospective image and user prompt output the final answer.
\begin{table}[h]
\centering
\caption{Generation study of truncation strategy on LLaVA \cite{LLaVA}, Intern-VL \cite{internvl}, Qwen \cite{bai2023qwen}, InstructBLIP2 \cite{instructblip} and shikra \cite{shikra}. The experimental results prove that this phenomenon of information flow convergence in the shallow layer and dispersion in the deep layer has widely existed. The i-f cliff layer represents the information flow cliff layer.}
\label{tab:prompts-llama3}
\scalebox{0.7}{
\begin{tabular}{@{}llccc@{}}
\toprule
Model & Dataset & Metric & Baseline & i-f cliff layer \\ 
\midrule
\multirow{5}{*}{LLaVA1.5 } 
& \multirow{1}{*}{POPE} & Acc $\uparrow$ & 84.70  & \textbf{85.51} \\ 
& \multirow{1}{*}{POPE} & F1-score $\uparrow$ & 85.50 & \textbf{85.99} \\ 
& \multirow{1}{*}{SQA} &Acc$\uparrow$ & 65.00 & \textbf{66.48} \\ 
& \multirow{1}{*}{TextVQA} & Acc $\uparrow$ & 59.34 & \textbf{60.02} \\ 
& \multirow{1}{*}{CHAIR} & CHAIR$_{s}$ $\downarrow$ & 13.80 & \textbf{13.80} \\ 
\midrule
\multirow{5}{*}{Qwen-VL } 
& \multirow{1}{*}{POPE} & Acc $\uparrow$ & 80.81 & \textbf{81.13} \\ 
& \multirow{1}{*}{POPE} & F1-score $\uparrow$ & 77.29 & \textbf{77.82} \\ 
& \multirow{1}{*}{SQA} & Acc$\uparrow$ & 67.12 & \textbf{67.67} \\ 
& \multirow{1}{*}{TextVQA} & Acc $\uparrow$ & 60.60 & \textbf{60.73} \\ 
& \multirow{1}{*}{CHAIR} & CHAIR$_{i}$ $\uparrow$ & 12.50 & \textbf{12.50} \\ 
\midrule
\multirow{5}{*}{Instruct-Blip2 } 
& \multirow{1}{*}{POPE} & Acc $\uparrow$ & 84.85 & \textbf{85.44} \\ 
& \multirow{1}{*}{POPE} & F1-score $\uparrow$ & 85.40 & \textbf{85.40} \\ 
& \multirow{1}{*}{SQA} & Acc$\uparrow$ & 60.50 & \textbf{60.50} \\ 
& \multirow{1}{*}{TextVQA} & Acc $\uparrow$ & 50.10 & \textbf{50.62} \\ 
& \multirow{1}{*}{CHAIR} & CHAIR$_{s}$ $\downarrow$  & 24.50 & \textbf{22.80} \\ 
\midrule
\multirow{5}{*}{Shikra } 
& \multirow{1}{*}{POPE} & Acc $\uparrow$ & 75.41 & \textbf{76.34} \\ 
& \multirow{1}{*}{POPE} & F1-score $\uparrow$ & 78.52 & \textbf{79.16} \\ 
& \multirow{1}{*}{SQA} & Acc$\uparrow$ & 45.80 & \textbf{46.70} \\ 
& \multirow{1}{*}{TextVQA} & Acc $\uparrow$ & - & - \\ 
& \multirow{1}{*}{CHAIR} &CHAIR$_{s}$ $\downarrow$ & 12.50 & 12.50 \\ 
\midrule
\multirow{5}{*}{Intern-VL } 
& \multirow{1}{*}{POPE} & Acc $\uparrow$ & 94.01 & \textbf{94.58} \\ 
& \multirow{1}{*}{POPE} & F1-score $\uparrow$ & 86.31 & \textbf{87.21} \\ 
& \multirow{1}{*}{SQA} & Acc$\uparrow$ & 66.20 & \textbf{66.85} \\ 
& \multirow{1}{*}{TextVQA} & Acc $\uparrow$ & 57.00 & \textbf{58.18} \\ 
& \multirow{1}{*}{CHAIR} &  CHAIR$_{s}$ $\downarrow$&12.90 & 12.90 \\ 
\midrule
\multirow{5}{*}{LLaVA1.6 } 
& \multirow{1}{*}{POPE} & Acc $\uparrow$ & - & - \\ 
& \multirow{1}{*}{POPE} & F1-score $\uparrow$ &86.50  & \textbf{86.99} \\ 
& \multirow{1}{*}{SQA} & Acc$\uparrow$ & 70.11 & \textbf{70.74} \\ 
& \multirow{1}{*}{TextVQA} & Acc $\uparrow$ & 64.9 & \textbf{65.23} \\ 
& \multirow{1}{*}{CHAIR} & CHAIR$_{i}$ $\uparrow$ & - & - \\ 
\bottomrule
\end{tabular}
}
\label{table1}
\end{table}





\subsection{Truncation Experiment in Different LVLMs}

As shown in Table \ref{table1}, to verify the generalization of the information flow convergence phenomenon across several models, we conduct experiments on Qwen-VL \cite{bai2023qwen}, LLaVA1.5 \cite{LLaVA}, Intern-VL \cite{internvl}, Shikra \cite{shikra}, InstructBlip \cite{instructblip} and LLaVA1.6 \cite{llava1.6}. Current LVLMs (Large Vision-Language Models) generally process images through a CLIP model, project them using various projectors, and then combine them with LLMs (Large Language Models) for fine-tuning. We speculate that the observed convergence of information flow in shallow layers is related to two factors: (1) The convergence is related to the multimodal paradigm. Different projectors, such as Linear, MLP, Cross-attention, and Q-former, map images to tokens in distinct ways. (2) This convergence is also related to the nature of LLMs. Since LVLMs are fundamentally LLMs, images are converted into tokens and aligned with text before being processed by the LLM. Thus, this pattern may be inherited from the underlying LLM structure.

To explore further, we conduct generalization experiments with 3 types of projectors. We select various models: LLaVA (Projector: MLP), Qwen-VL (Projector: Cross-attention), Intern-VL (Projector: MLP), InstructBlip (Projector: MLP), and Shikra (Projector: Linear). As shown in Table \ref{table1}, all models exhibit a consistent pattern of shallow layer convergence and deep layer dispersion. Moreover, the performance after deep layer truncation is equal to or surpasses the baseline, suggesting that shallow layer convergence is independent of the projector type and is instead related to the LLM itself.

The SFT+LLM paradigm does not fundamentally alter the reasoning process of the LLM, as noted in references \cite{label-words,jieshi1,jieshi2}, which may come from that shallow layers extract basic features and patterns from the input data, learning essential semantic information. The input takes the form of a `Prompt + Question', where the information flow initially focuses on specific anchor tokens at a shallow layer. At these shallow layers, the model is processing semantic understanding, while at a deeper layer, it engages in more complex reasoning and thinking. This process is similar in Large Vision Language Models (LVLMs), where the prompt structure follows the format `<image> Prompt + Question'. At the early stage, the model focuses on understanding the image and the prompt, and the middle layers combine these basic features to form more complex representations, while the deep layers perform high-level modeling and reasoning, integrating information from earlier layers to generate the final understanding and decision of the task.

\section{Conclusion}

In this paper, we employ LLaVA-CAM and attention score to visualize the information flow of reasoning processes layer-by-layer within LVLMs. Our analysis of several models shows the same pattern that information flow converges in the shallow layers and diverges in the deep layers. Comprehensive experiments using truncation strategies across various large-vision language models for VQA and image captioning tasks further validate our findings, confirming a consistent pattern of information flow convergence in the relevant layers across different models. These insights contribute to a deeper understanding of LVLMs and their intrinsic functioning, particularly in the context of complex reasoning tasks.

\section{limitations}
Apart from the contributions introduced in the main paper, there are limitations of this work:
(1) Due to resource limitations, we explored the information flow merely on the 7B model and have not yet tried the 13B or larger models. 
(2) Given the black-box nature of multimodal large models, some of the arguments in this paper are drawn from experiments and lean slightly towards practical observations. Other arguments, such as the assumption that 32 layers are retrospective layers, are speculative and remain unverified. (3) LLaVA-CAM is primarily used to provide a visual interpretation by visualizing the gradient information at a particular layer that shows the image regions the model focuses on when making predictions. While this is useful for understanding the interaction mechanisms between image and text, it may not be sufficient to fully capture textual components' complex interaction and reasoning processes. Attention scores can reveal what the model focuses on when processing input, but they are static and do not fully reflect the dynamics of the model throughout the reasoning process.
(4) Our demonstration of redundancy in image features and aggregation of image tokens over salient tokens through truncation techniques also reveals an important issue: even with advanced visualization and interpretation methods, it remains a challenge to effectively handle and reduce information redundancy to improve model efficiency and interpretability.
\bibliography{custom}

\newpage
\appendix

\section{Appendix}
\label{sec:appendix}
\subsection{LLaVA-CAM Structure}

\begin{figure}[h]
\centerline{\includegraphics[scale=0.21]{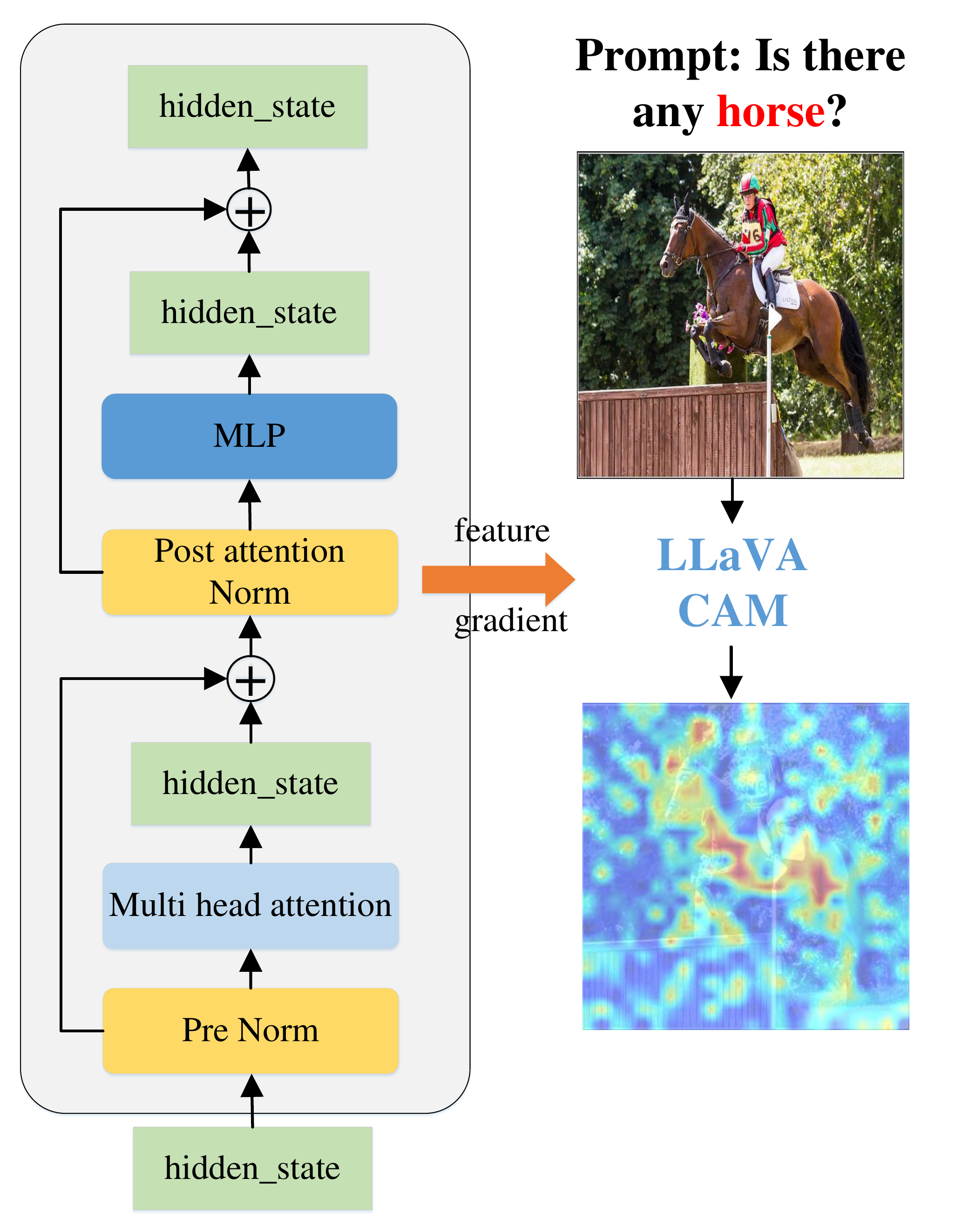}}
\caption{LLaVA-CAM selects features after the post-attention layer normalization to reduce numerical instability and produce clearer heatmaps.}
\label{Smooth-CAM-llava}
\end{figure}

\begin{figure}[h]
\centerline{\includegraphics[scale=0.16]{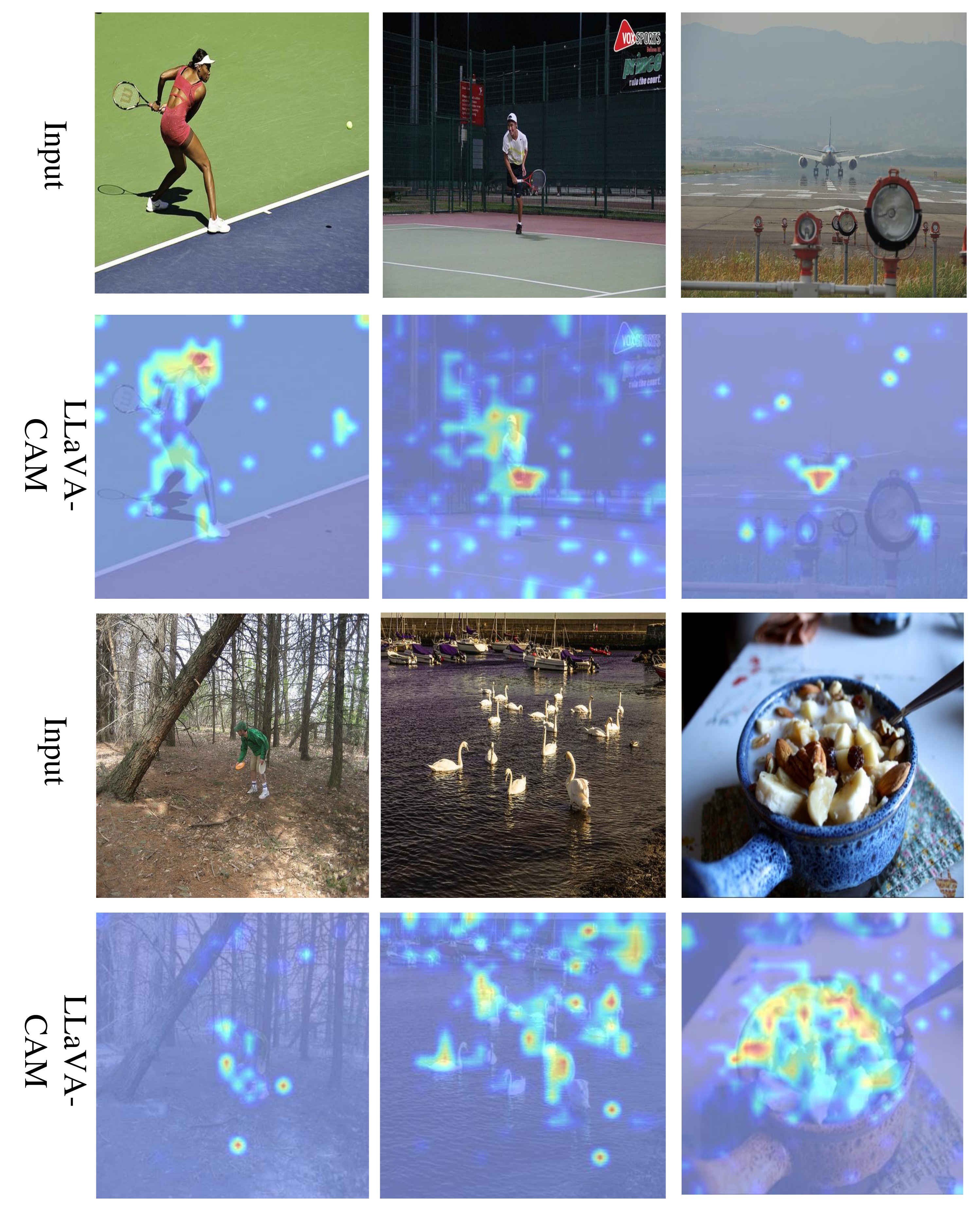}}
\caption{The LLaVA-CAM results.}
\label{more-cam}
\vspace{-0.4cm}
\end{figure}

The LLaVA-CAM is shown in Figure~\ref{Smooth-CAM-llava}, and more results are shown in Figure~\ref{more-cam}. The results show that LLaVA-CAM visualization is clearest after the post-attention layer normalization. This is because the post-attention-layer norm stabilizes the output of the self-attention mechanism by normalizing it. This normalization process helps stabilize the feature distribution, reducing the impact of numerical instability and making the features more consistent and reliable, which is crucial for generating high-quality heat maps. In contrast, feature maps from other positions may lack the same level of normalization or feature expression, resulting in lower-quality heatmaps and difficulty in accurately locating the target object. The feature maps after the attention mechanism contain rich contextual information.

\subsection{Redundancy Truncation Experiment in Shallow Layers}

To explore the saliency and information flow of image tokens in the shallow layer, we designed an early truncation strategy. We ranked the attention scores and selected the top K image tokens to pass through, simplifying the inference process by focusing on these key tokens. We applied this attention-score truncation strategy with different values of K: 0.

\begin{figure}[h]
\centerline{\includegraphics[scale=0.3]{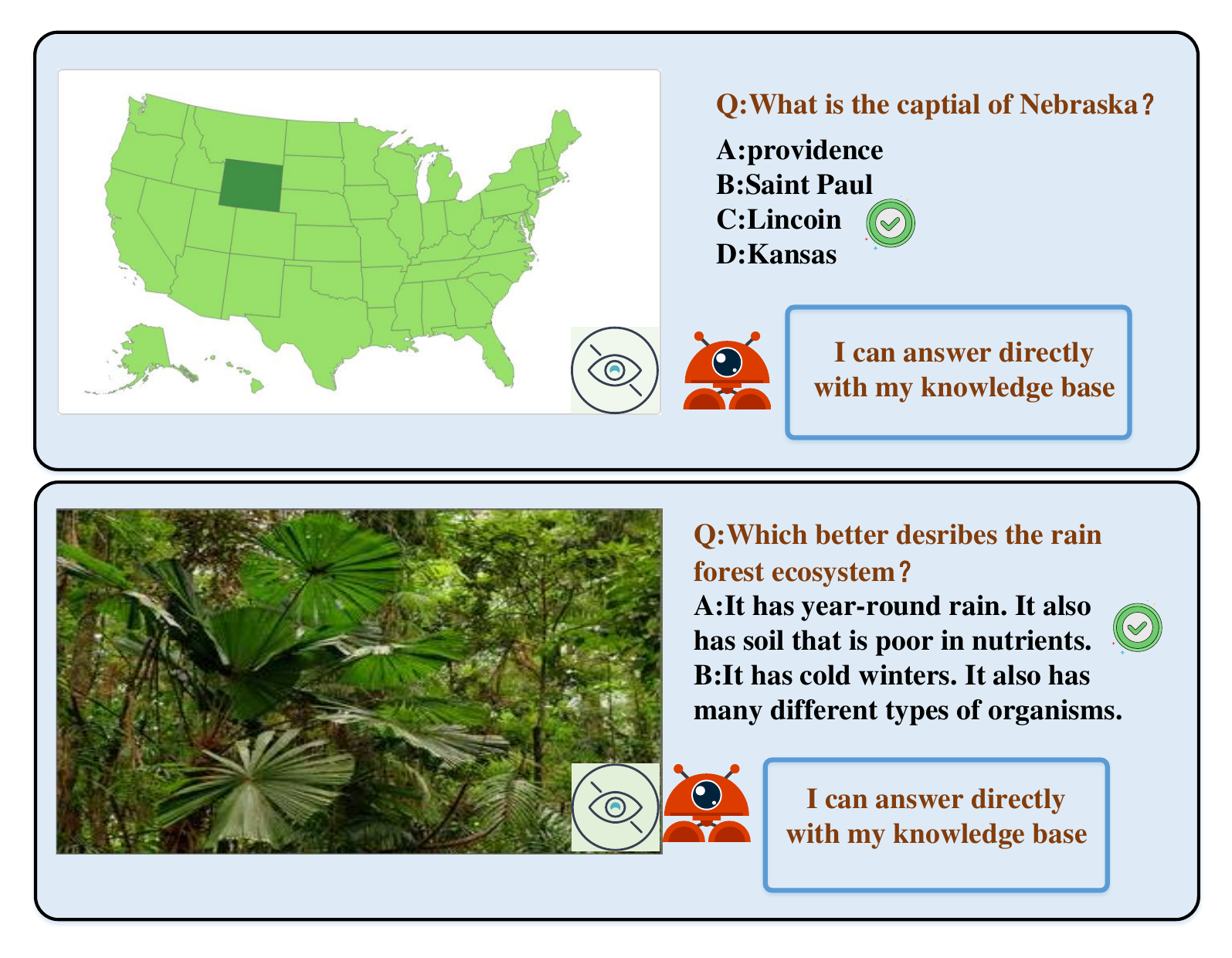}}
\caption{It shows the case lacks visual dependency.}
\label{sqa-noimage}
\vspace{-0.4cm}
\end{figure}

When the layer=1, there is no image token passed to the LLM, while the LLM can surprisingly achieve an accuracy of 47.5 $\%$, which suggests that in some scenarios the LLM relies only on the text to answer the questions on their own through prior knowledge, such as \emph{`What is the capital of Nebraska?’} in Figure~\ref{sqa-noimage}. Therefore, LLM does not attend to graphical information when LLM's knowledge can directly answer a question.

\end{document}